\documentclass[lettersize,journal]{IEEEtran}
\usepackage{amsmath,amsfonts}
\usepackage{algorithmic}
\usepackage{array}
\usepackage[caption=false,font=normalsize,labelfont=sf,textfont=sf]{subfig}
\usepackage{textcomp}
\usepackage{stfloats}
\usepackage{url}
\usepackage{verbatim}
\usepackage{graphicx}
\usepackage{cite}
\usepackage[T1]{fontenc}
\usepackage{xcolor}
\usepackage{makecell} 
\usepackage{multirow}  
\usepackage{pgfplots}
\usepackage[ruled,vlined,linesnumbered]{algorithm2e}
\definecolor{myteal}{RGB}{0,128,128}
\definecolor{myorange}{RGB}{230,120,20}
\usepackage{booktabs}
\begin{document}

\title{Recursive Hierarchical Feature Refinement for Human Pose Estimation}


\author{{Shibang Liu, Xuyang Li, Xuemei Xie,~\IEEEmembership{Senior Member,~IEEE}, Yi Luo}
\thanks{Manuscript received April X, X; revised August X, X. This work was supported in part by National Science and Technology Major Project-Mobile Information Networks under Grant No.2024ZD1300700 and in part by the Natural Science Foundation (NSF) of China under Grant 62293483. (\emph{Corresponding author: Xuyang Li.})}
\thanks{Shibang Liu is with the School of Artificial Intelligence, Xidian University, 
	Xi'an 710071, China (e-mail: shibangliu@stu.xidian.edu.cn).}
\thanks{Xuyang Li is with the Guangzhou institute of technology, Xidian University, Guangzhou, 510555, Guangdong, China (e-mail: lixuyang01@xidian.edu.cn)}
\thanks{Xuemei Xie is with the School of Artificial Intelligence, Xidian University, Xi'an 710071, China, Guangzhou Institute of Technology, Xidian University, Guangzhou 510530, China and Pazhou LAB, Huangpu, Guangzhou 510555, China (e-mail: xmxie@mail.xidian.edu.cn).}
\thanks{Yiluo is with the Guangzhou institute of technology, Xidian University, Guangzhou, 510555, Guangdong, China (e-mail: yiluo@stu.xidian.edu.cn).}
}

\markboth{Journal of \LaTeX\ Class Files,~Vol.~14, No.~8, August~2021}%
{Shell \MakeLowercase{\textit{et al.}}: A Sample Article Using IEEEtran.cls for IEEE Journals}

\IEEEpubid{0000--0000/00\$00.00~\copyright~2021 IEEE}

\maketitle

\begin{abstract}
Human body structures exhibit recursive compositionality, providing rich hierarchical structure and context relations priors for Human Pose Estimation (HPE). Existing methods typically model these priors using specially designed network architectures (e.g., hierarchical branches), introducing additional parameter overhead. Inspired by this property, we propose RHFR (Recursive Hierarchical Feature Refinement), a plug-and-play module that refines features through recursive top-down decomposition and bottom-up composition.
Specifically, RHFR organizes feature representations into a hierarchical tree through top-down decomposition. During bottom-up composition, a novel Node Attention (NA) captures context relations among nodes to enhance their representations, which are then recursively aggregated toward the root. This process progressively propagates context information across feature hierarchies, facilitating occluded-region inference and improving joint localization. Moreover, the recursively generated nodes also can be directly supervised by human structural priors, enabling explicit hierarchical structure and context relation modeling without additional network branches, thereby reducing parameter overhead.
Extensive experiments demonstrate that RHFR consistently improves performance across various pose estimation frameworks while enabling efficient explicit human structure modeling. The source code is available at \url{https://github.com/lushbng/RHFR}.

\end{abstract}


\section{Introduction}
\label{sec:intro}

\IEEEPARstart{T}{he} main task of 2D Human Pose Estimation (HPE) is to localize human body joints in images to infer overall body configuration. It has been widely applied in security surveillance, behavior analysis, rehabilitation monitoring, and other fields. In general, HPE methods follow two paradigms: top-down (detect-then-estimate)~\cite{KnowledgeGuide,HumanBodyAwareFeature,Hybrid} and bottom-up (group-joints)~\cite{GroupingByCenter,BoundingBoxLSTM}. In this work, we adopt the top-down paradigm for single-person pose estimation.

In HPE, a key issue lies in how to effectively leverage inherent human structural priors to improve joint localization accuracy. The human body exhibits a recursive compositional hierarchy, where joints form local body parts and these parts further compose the full body, resulting in rich hierarchical structures and contextual dependencies. Meanwhile, human visual perception also relies on hierarchical structure and context relations to understa body configurations, which can be formally described using parse graphs~\cite{GrammarOfImages}.

 \begin{figure}[t]
	\centering
	
	\includegraphics[width=1\linewidth]{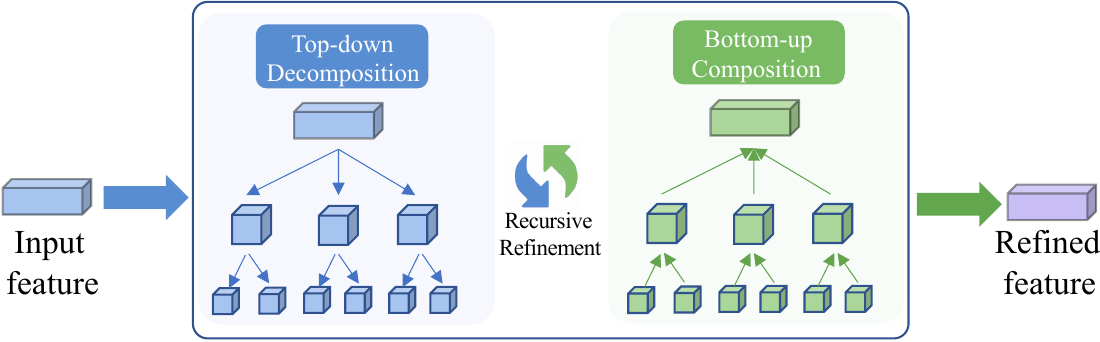}
	
	\caption{Overview of RHFR. The proposed RHFR performs feature refinement through top-down decomposition and bottom-up composition. }
	\label{fig1}
\end{figure}

Existing methods attempt to incorporate these priors from two main perspectives. Hierarchical structure modeling approaches~\cite{DeepFullyConnected,DeeplyLearnedCompositionalModels,EHPE} explicitly encode hierarchical body structures to guide feature learning and inference. Context relations modeling methods~\cite{SpatialContextual,TokenPose,PCTPose,Graphpcnn,DGN} focus on capturing dependencies among joints or body regions to enhance feature interactions. For example, PCTPose~\cite{PCTPose} models contextual relations among pose tokens, while Graph-PCNN~\cite{Graphpcnn} and DGN~\cite{DGN} model joint interactions through graph convolution. PGBS~\cite{PGBS} integrates hierarchical structure and context relations modeling into a unified framework. However, PGBS rely on designed additional hierarchical branch architectures, which introduce additional parameter overhead.\IEEEpubidadjcol

To address these limitations, inspired by the recursive compositional structure of the human body, we propose RHFR (Recursive Hierarchical Feature Refinement), a plug-and-play module that performs recursive feature refinement in feature space, as illustrated in Fig.~\ref{fig1}. RHFR adopts a top-down decomposition and bottom-up composition strategy, progressively decomposing features into implicit sub-feature components and aggregating them across hierarchical levels. Through this process, RHFR integrates global semantics with local details, while a novel Node Attention (NA) establishes context relations among nodes during bottom-up refinement. Moreover, the generated nodes can be supervised by human structural priors to efficiently enable explicit structural modeling. We construct a hierarchical network by cascading RHFR modules. Specifically, the first RHFR module is supervised at the body-part level, guiding NA to learn context relations among body parts. The second module is then supervised at the joint level to capture context relations among joints. This coarse-to-fine hierarchical learning strategy progressively incorporates structural context at different levels, leading to more robust joint localization~\cite{GrammarOfImages}. Compared with the he same-category explicit structured modeling method PGBS~\cite{PGBS}, our network achieves competitive performance with a simpler architecture and fewer parameters.
In summary, our contributions are as follows:
\begin{itemize}
	\item We propose RHFR, a plug-and-play module that performs recursive feature refinement through top-down decomposition and bottom-up composition in feature space.
	\item  We propose a NA mechanism that models context relations among nodes for effective recursive contextual reasoning.
	\item Our hierarchical network shows that RHFR supports structural supervision for explicit modeling human body structure with lower parameter overhead compared with PGBS~\cite{PGBS}.
\end{itemize}

 \begin{figure*}[h]
	\centering
	
	\includegraphics[width=0.8\linewidth]{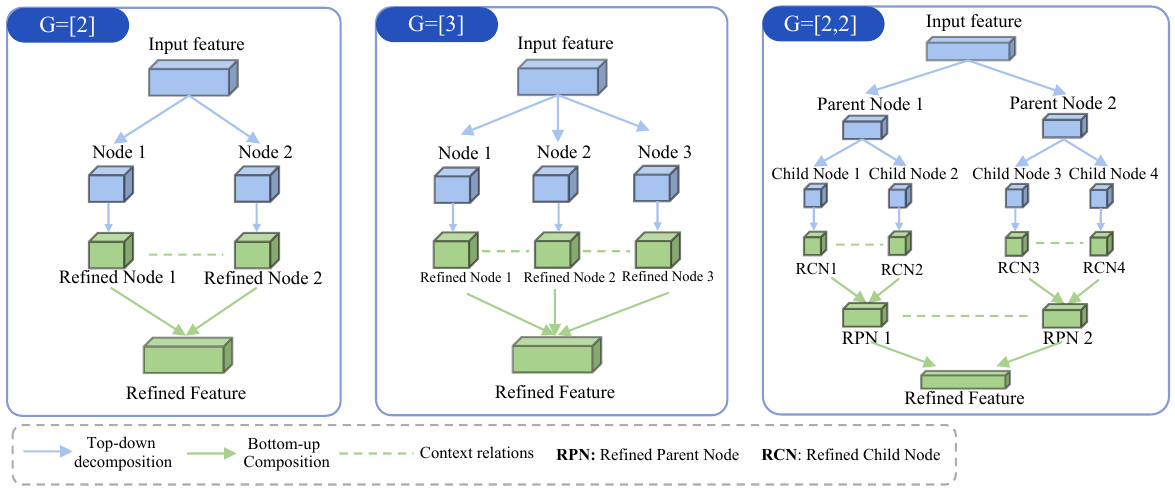}
	
	\caption{The examples of RHFR and its expansion. RHFR supports two types of structural expansion: breadth expansion adds more child nodes to a parent at the same level, increasing the number of nodes in that layer, and depth expansion, which increases the number of hierarchical levels}
	\label{fig2}
\end{figure*}

\begin{figure*}[h]
	\centering
	
	\includegraphics[width=1\linewidth]{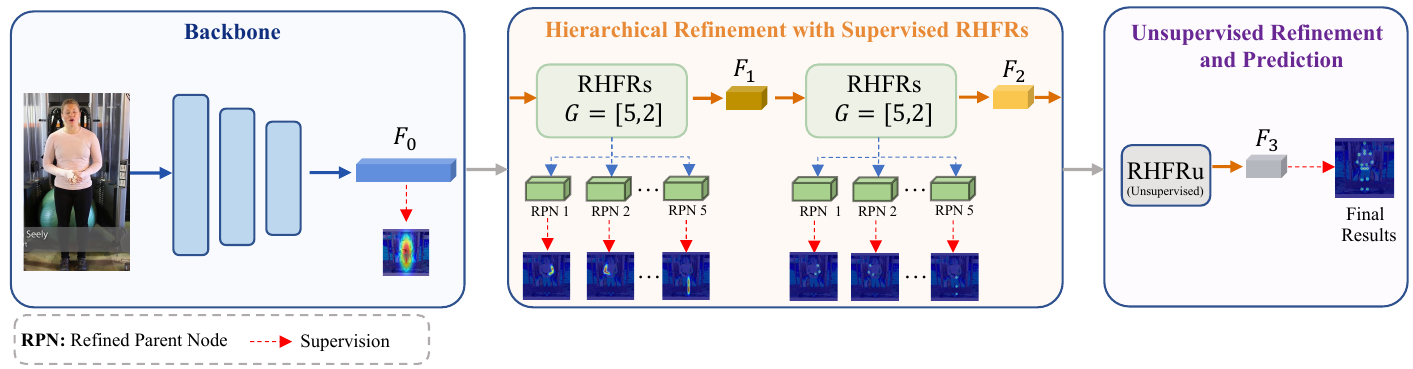}
	
\caption{Overview of our hierarchical network. Human representations are organized into three levels: full body, body parts, and joints. Prediction heads are omitted for clarity. Each Refined Parent Node (RPN) feature map produced by RHFR$_\text{s}$ (see Fig.~\ref{fig2}) is fed into a lightweight 2D Conv head for heatmap prediction and loss computation.}
	\label{fig3}
\end{figure*}

\section{Related Work}
\label{sec:rel}

{\bf Hierarchical Modeling in HPE.}
Modeling the human body as a hierarchical composition has long been a fundamental idea in human pose estimation~\cite{GrammarOfImages}. Early works often represent the human body using tree-structured or compositional models, where body parts are recursively assembled from joints to form higher-level semantic units~\cite{MultiScaleStructureAware,DeeplyLearnedCompositionalModels,DeepFullyConnected}. Such methods explicitly encode structural priors to improve part localization, but usually rely on hand-crafted or explicitly defined hierarchical designs, limiting their flexibility in deep networks.

{\bf Context Modeling for Joint.}
Modeling contextual dependencies among human joints is a key factor in improving pose estimation accuracy. Early CNN-based methods~\cite{Hourglass,SimpleBaseline} implicitly capture local context through convolutional operations with limited receptive fields, where hierarchical stacking gradually enlarges the spatial dependency range. However, such local modeling is still insufficient to capture long-range interactions between distant joints.
To address this limitation, recent Transformer-based approaches, such as ViTPose~\cite{Vitpose} TokenPose~\cite{TokenPose}, and GTPT~\cite{GTPT}, leverage self-attention mechanisms to model global relationships among all joints, enabling more effective long-range dependency reasoning. In addition, graph convolutional methods~\cite{Graphpcnn,DGN} explicitly construct joint interaction graphs to propagate contextual information across the body structure.

{\bf Unified Structural Representation.}
To jointly model hierarchical structure and context relations, parse graph-based methods have been introduced in human vision tasks~\cite{GrammarOfImages,PGBS}. By representing the human body as a graph with multi-level semantic nodes, these methods enable structured reasoning across different granularity levels. For example, PGBS~\cite{PGBS} explicitly constructs hierarchical representations and models interactions across levels. However, such approaches typically depend on carefully designed network structures and introduce additional architectural complexity.

\section{Method}
\label{sec:meth}

\subsection{Motivation}

Human body structures naturally exhibit a recursive compositional organization, forming rich hierarchical structures and context relations that are important priors for HPE. To better illustrate this property, we introduce the parse graph formulation~\cite{GrammarOfImages} as a conceptual abstraction of human body structure.

\textbf{Parse Graph.} A parse graph captures two aspects~\cite{GrammarOfImages}: (i) the hierarchical structure of the human body, and (ii) contextual relationships among body parts. It is represented as a 4-tuple $(\mathcal{V}, \mathcal{E}, \psi^{\text{and}}, \psi^{\text{leaf}})$, where $(\mathcal{V}, \mathcal{E})$ defines the hierarchical structure and $(\psi^{\text{and}}, \psi^{\text{leaf}})$ denote potential functions. Each node $u \in \mathcal{V}$ is associated with a state variable $\mathbf{s}_u = \{x_u, y_u\}$, where $x_u$ and $y_u$ represent its position and type, respectively. The corresponding energy-based score function is defined as~\cite{PGBS}:
\begin{equation}
	\label{scorefunction}
	F(\boldsymbol{\Omega}) =
	\sum_{u \in \mathcal{V}_{\text{leaf}}}
	\psi^{\text{leaf}}_u(\mathbf{s}_u, I)
	+
	\sum_{u \in \mathcal{V}_{\text{and}}}
	\psi^{\text{and}}_u
	\left(
	\mathbf{s}_u,
	\{\mathbf{s}_v\}_{v \in \mathcal{C}(u)}
	\right),
\end{equation}
where $\mathcal{V}_{\text{leaf}}$ and $\mathcal{V}_{\text{and}}$ denote leaf and non-leaf nodes, respectively, $\mathcal{C}(u)$ denotes the children of node $u$, $\boldsymbol{\Omega}$ denotes the collection of node states, and $I$ is the input image. The energy-based formulation captures local appearance and hierarchical compositional dependencies. Based on this formulation, PGBS~\cite{PGBS} performs bottom-up activation and top-down refinement to infer the optimal configuration of node states.

Inspired by the recursive compositional property and contextual reasoning of parse graphs, we design RHFR that performs recursive hierarchical reasoning in feature space.

\subsection{RHFR}

RHFR, a plug-and-play module that performs recursive hierarchical feature refinement directly in feature space.
Instead of performing parse graph-based inference, 
RHFR reformulates hierarchical reasoning as a learnable message passing process, where a novel  NA mechanism implicitly models context relations among nodes.
Specifically, for a feature map, the top-down decomposition and bottom-up composition processes are unified into a recursive feature refinement framework.

RHFR is defined as
\begin{align}
	\text{RHFR}
	\left(
	{f}_v^{},
	G
	\right)
	=	{f}_v^{}+
	\Phi_{\text{BU}}
	\left(
	\Phi_{\text{TD}}
\left(
{f}_v,
G
\right)
	\right),
\end{align}
where ${f}_v$ denotes the feature representation of node $v$, and $G$ defined in Eq.~\ref{G} is a hierarchical structure descriptor specifying the number of levels and child nodes at each level.

Starting from the root node ${f}_v$, $\Phi_{\text{TD}}(\cdot)$ recursively decomposes the input feature according to the hierarchical descriptor $G$. The decomposition proceeds level by level until the predefined leaf level is reached, resulting in a hierarchical parse tree $\mathcal{T}$. $\Phi_{\text{BU}}(\mathcal{T})$ then performs recursive bottom-up composition from leaf nodes to the root, progressively integrating contextual information across different hierarchy levels to obtain the refined feature representation. The refined feature is then added to the input through a residual connection.

\textbf{Top-down Decomposition:}

The hierarchical structure is defined by
\begin{align}
	\label{G}
	G=[g_d,g_{d-1},\ldots,g_1],
\end{align}
where $g_i$ denotes the number of child nodes generated from each node at hierarchy level $i\in \{0,\cdots,d-1\}$, and $d$ represents the hierarchy depth.

Given the input feature representation
$F\in\mathbb{R}^{L\times C}$,
RHFR recursively decomposes the feature representation along the channel dimension to construct a hierarchical feature tree. The input feature is regarded as the root node at the top hierarchy level:
\begin{align}
	P_d^1=F,
\end{align}
where the root level contains only one node, i.e.,
\begin{align}
	N_d=1.
\end{align}

Let $\mathcal{V}_i$ denote the set of nodes at hierarchy level $i$:
\begin{align}
	\mathcal{V}_i
	=
	\{
	P_i^1,
	P_i^2,
	\dots,
	P_i^{N_i}
	\},
\end{align}
where $N_i$ represents the number of nodes at level $i$, $P^j_i \in \mathbb{R}^{L\times \frac{C}{N_i}},j\in \{1,\cdots,N_i\}$.

Starting from the root node, RHFR first decomposes it into $g_d$ child nodes at hierarchy level $d-1$:
\begin{align}
	\mathcal{V}_{d-1}
	=
	\Phi_{\mathrm{split}}
	(
	P_d^1
	).
\end{align}

For subsequent hierarchy decomposition, each node $P_i^j\in\mathcal{V}_i$ is recursively decomposed into $g_i$ child nodes:
\begin{align}
	\{
	P_{i-1}^{j,1},
	P_{i-1}^{j,2},
	\ldots,
	P_{i-1}^{j,g_i}
	\}
	=
	\Phi_{\mathrm{split}}
	(
	P_i^j
	),
\end{align}
where $i\in\{1,\ldots,d-1\}$ and $\Phi_{\mathrm{split}}(\cdot)$ partitions the channel dimension of the parent feature into $g_i$ feature subspaces, with each child node inheriting the corresponding feature information from its parent.

By applying this operation to all nodes at hierarchy level $i$, the node set at the next hierarchy level is obtained as
\begin{align}
	\mathcal{V}_{i-1}
	=
	\bigcup_{j=1}^{N_i}
	\Phi_{\mathrm{split}}
	(
	P_i^j
	).
\end{align}
Consequently, the number of nodes at the next hierarchy level is given by
\begin{align}
	N_{i-1}
	=
	N_i\times g_i.
\end{align}

By recursively applying the decomposition operation from the root node to the predefined leaf level, the hierarchical feature tree is progressively constructed:
\begin{align}
	\mathcal{T}
	=
	\Phi_{\mathrm{TD}}
	(F,G).
\end{align}

\textbf{Bottom-up Composition:}

Given the hierarchical feature tree $\mathcal{T}$ constructed by the top-down decomposition, the bottom-up composition recursively updates node representations from the leaf level to the root level.

Different from conventional self-attention that operates on spatial tokens, we propose a novel Node Attention (NA) mechanism to model context relations among sibling nodes, i.e., child nodes belonging to the same parent node.

Starting from the leaf hierarchy $\mathcal{V}_0$, the bottom-up process updates each parent node by first performing contextual reasoning among its child nodes and then merging their enhanced representations.

Specifically, for each parent node $P_{i+1}^j$, its corresponding child nodes are denoted as
${P_i^{j,1},P_i^{j,2},\ldots,P_i^{j,g_{i+1}}}$.
NA is independently applied to these child nodes to capture their contextual dependencies:
\begin{align}
	{
		\hat P_i^{j,1},
		\hat P_i^{j,2},
		\ldots,
		\hat P_i^{j,g_{i+1}}
	}
	=
	\mathrm{NA}
	\left(
	{
		P_i^{j,1},
		P_i^{j,2},
		\ldots,
		P_i^{j,g_{i+1}}
	}
	\right),
\end{align}
where $\hat P_i^{j,k}$ denotes the enhanced representation of the $k$-th child node under the $j$-th parent node.

The enhanced child nodes are then merged through channel-wise concatenation to update their corresponding parent node:
\begin{align}
	 P_{i+1}^{j}=
	\hat P_{i+1}^{j}
	=
	\mathrm{Concat}
	(
	\hat P_i^{j,1},
	\hat P_i^{j,2},
	\ldots,
	\hat P_i^{j,g_{i+1}}
	),
\end{align}
where $\hat P_{i+1}^{j}$ denotes the updated parent representation, and the concatenation operation restores its original channel dimension.

After all parent nodes at the current hierarchy level have been updated, their refined representations are used as child nodes for the next recursive composition step. This process is progressively performed from the leaf level to the root level, which $i$ increasing from 0 to d-1:
\begin{align}
	\mathcal{V}_{i+1}=\hat {\mathcal{V}}_{i+1}
	=
\left\{
\mathrm{Concat}
\left(
\mathrm{NA}
\left(
\left\{
P_i^{j,k}
\right\}_{k=1}^{g_{i+1}}
\right)
\right)
\right\}_{j=1}^{N_{i+1}}.
\end{align}

Finally, the root representation is obtained after the last merging operation:
\begin{align}
	F_{\mathrm{root}}
	=	 P_d^1=
	\hat P_d^1
\end{align}

Since the channel-wise decomposition and subsequent composition 
may disrupt the original channel structure, a residual connection
is introduced to preserve the original feature representation.
The refined root feature is therefore added to the input feature:
\begin{align}
	F'
	=
	F+F_{\mathrm{root}}.
\end{align}

\subsection{Node Attention}

\begin{figure}[!t]
	\centering
	\subfloat[]{\includegraphics[width=0.6in]{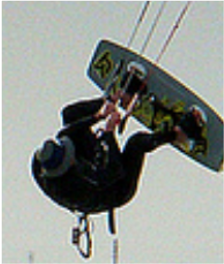}%
		\label{figrhfr-a}}
	\hspace{0.3cm}
	\subfloat[]{\includegraphics[width=0.6in]{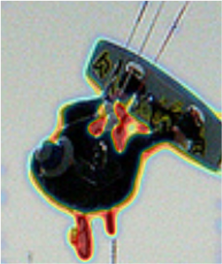}%
		\label{figrhfr-b}}
	\hspace{0.1cm}
	\subfloat[]{\includegraphics[width=0.6in]{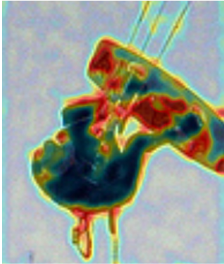}%
		\label{figrhfr-c}}
	\hspace{0.1cm}
	\subfloat[]{\includegraphics[width=0.6in]{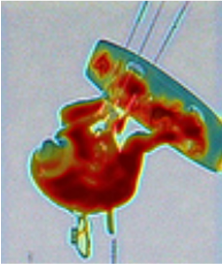}%
		\label{figrhfr-d}}
	\hspace{0.1cm}
	\subfloat[]{\includegraphics[width=0.6in]{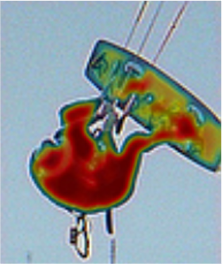}%
		\label{figrhfr-e}}
	\caption{Visualization of feature response intensities mapped onto the input image at different hierarchical stages of RHFR. (a) Original input image; (b) RHFR input feature map response; (c) first refined parent node feature map response; (d) second refined parent node feature map response; (e) renfined root node feature map response.}
	\label{figrhfr}
\end{figure}

Unlike conventional spatial self-attention over spatial tokens, NA dynamically models context relations among sibling feature maps sharing the same parent, enabling efficient information exchange across feature subspaces at each spatial location.
 Specifically, NA performs contextual reasoning among sibling nodes belonging to the same parent node, enabling information exchange across different feature subspaces. This interaction enhances human-related feature responses and promotes more concentrated feature representations around the human body, thereby improving pose estimation performance (see Fig.~\ref{figrhfr}). Moreover, by restricting attention to sibling nodes rather than all spatial tokens, NA substantially reduces the attention computation.

For each parent node $P_{i+1}^j\in \mathbb{R}^{L\times \frac{C}{N_{i+1}}}$, its corresponding child nodes are denoted as
$\{P_i^{j,1},P_i^{j,2},\ldots,P_i^{j,g_{i+1}}\}$, where each child node belongs to $\mathbb{R}^{L\times \frac{C}{N_i}}$.

To perform node contextual reasoning, the child node features are stacked along a new dimension:
\begin{align}
	X_i
	=
	\mathrm{Stack}
	(
	\{P_i^{j,1},P_i^{j,2},\ldots,P_i^{j,g_{i+1}}\}
	),
\end{align}
where $X_i\in\mathbb{R}^{L\times g_{i+1}\times C_i}$ and $C_i=\frac{C}{N_i}$.

The contextual relations among child nodes are then modeled as
\begin{align}
	A_i
	=
	\mathrm{Softmax}
	\left(
	\frac{
		X_iX_i^{T}
	}
	{\sqrt{C_i}}
	\right),
\end{align}
where $A_i\in\mathbb{R}^{L\times g_{i+1}\times g_{i+1}}$ denotes the node attention matrix. For each spatial token, $A_i$ captures the contextual dependencies relations among child nodes.

The node-enhanced feature representation is obtained by aggregating contextual dependencies relations from child nodes:
\begin{align}
	X_i^{\mathrm{NA}}
	=
	A_iX_i,
\end{align}
where $X_i^{\mathrm{NA}}\in\mathbb{R}^{L\times g_{i+1}\times C_i}$. Through this operation, each child node incorporates complementary information from its siblings, producing contextually enhanced feature representations. Finally, the enhanced child node features are concatenated along the channel dimension to reconstruct and update their corresponding parent node $P_{i+1}^j$. The overall procedure of RHFR is summarized in Algorithm~\ref{algorithm:RHFR}.

\subsection{Hierarchical Modeling Network}

To evaluate the capability of RHFR for efficient and explicit human structure modeling, we construct the hierarchical network using the backbone (HRNet-W32~\cite{Hrnet}) as same as PGBS~\cite{PGBS}, ensuring a fair comparison in terms of structural modeling performance and parameter overhead.

As illustrated in Fig.~\ref{fig3}, our network is built upon HRNet-W32~\cite{Hrnet}. The feature map $F_0 \in \mathbb{R}^{L\times C}$ is extracted from the highest resolution stage of the fourth stage and used as the input of RHFR.
Without introducing additional hierarchical branches or dedicated structural modules as in PGBS, we simply embed RHFR into the existing backbone and perform hierarchical reasoning directly in the hierarchical feature space.

As shown in the left and middle panels of Fig.~\ref{fig3}, RHFR is applied in a cascaded manner to progressively refine representations from body-level to part-level and then to joint-level.

\textbf{Hierarchical Supervision:}

As shown in the middle panel of Fig.~\ref{fig3}, in embeded the RHFR, the recursive decomposition and composition process naturally produces intermediate sub-features, which allows us to introduce structural supervision in a lightweight manner without additional architectural design.

Following the recursive decomposition and composition strategy illustrated in Fig.~\ref{fig2}, we adopt the same human body-part grouping strategy as PGBS for a fair comparison, dividing the body into five parts: two arms, two legs, and the torso. The first supervised RHFR$_\text{s}$ adopts $G=[5,2]$ to produce five parent node features expect root node. Supervision is then imposed on the refined parnet node $\{P_1^{1,k}\}_{k=1}^{5}$ using body-part heatmaps corresponding to two arms, two legs, and the torso, thereby introducing explicit structural priors into the learned representations.
In the second RHFR$_\text{s}$, the same decomposition strategy $G$ is applied, while supervision is switched from body-part to joint (e.g., conversion from left-arm supervision to left shoulder, left elbow, and left wrist joint supervision), enabling finer-grained structural learning.
All supervision signals are imposed on the refined parent node features after bottom-up composition.

As shown in the right panel of Fig.~\ref{fig3}, an additional RHFR$_\text{u}$ module is introduced before the final prediction to alleviate the constraints imposed by explicit hierarchical supervision on feature representations, allowing further adaptive refinement beyond structural priors.

\textbf{Design of Supervision Labels:}

Following PGBS~\cite{PGBS}, body heatmaps are generated using Gaussian kernels centered on the human body region, with kernel size proportional to body scale. Part heatmaps are generated at skeletal midpoints, with kernel size proportional to bone length. For example, the left leg heatmap includes Gaussian kernels at the midpoints of the hip--knee and knee--ankle segments. Joint heatmaps are generated using Gaussian kernels centered at the ground-truth joint locations.

\begin{algorithm}[t]
	\caption{RHFR}
	\label{algorithm:RHFR}
	
	\KwIn{
		Feature map $F \in \mathbb{R}^{L\times C}$,
		hierarchy descriptor $G=[g_d,\dots,g_1]$
	}
	\KwOut{
		Enhanced feature map $F'$
	}
	
	\tcc{Top-down decomposition}
	$P_d^1 \leftarrow F$\;
	$\mathcal{T}
	\leftarrow
	\Phi_{\mathrm{TD}}
	(P_d^1,G)$\;
	
	\tcc{Bottom-up composition}
	\For{$i=0$ \KwTo $d-1$}{
		
		\For{each parent node $P_{i+1}^j$}{
			
			$\{P_i^{j,1},\ldots,P_i^{j,g_{i+1}}\}
			\leftarrow
			\mathrm{Children}(P_{i+1}^j)$\;
			
			$\{\hat P_i^{j,1},\ldots,\hat P_i^{j,g_{i+1}}\}
			\leftarrow
			\mathrm{NA}
			(\{P_i^{j,1},\ldots,P_i^{j,g_{i+1}}\})$\;
			
			$\hat P_{i+1}^j
			\leftarrow
			\mathrm{Concat}
			(\hat P_i^{j,1},\ldots,\hat P_i^{j,g_{i+1}})$\;
			
			$P_{i+1}^j
			\leftarrow
			\hat P_{i+1}^j$\;
		}
	}
	
	\tcc{Residual refinement}
	$F' \leftarrow F+P_d^1$\;
	
	\Return{$F'$}\;
	
\end{algorithm}

\begin{table*}[htbp]
\caption{Performance of RHFR on different methods using a single RHFR module on the COCO validation set. $\dagger$ denotes our reimplemented results.}
	\label{tb1}
	\centering
		\begin{tabular}{l|c
				|l|l|l|c|l|c}	
			\hline
			\multicolumn{1}{c|}{Method} & \multicolumn{1}{c|}{RHFR} & \multicolumn{1}{c|}{Backbone} & \#Params &GFLOPs& Input size &  \multicolumn{1}{c|}{MAP} &  \multicolumn{1}{c}{MAR} \\ 
			\hline
			\multicolumn{7}{c}{Baselines} \\ 
			\hline
			Hourglass-52~\cite{Hourglass} &  \multicolumn{1}{c|}{-} & Hourglass-52 & 94.8M&28.7 & 256$\times$256 & 72.6 & 78.0 \\
						
			SWIN~\cite{SWIN} &  \multicolumn{1}{c|}{-}& swin\_b  & 93.0M &19.0& 256$\times$192 & 73.7&79.4\\ 
			
			ViTPose$\dagger$~\cite{Vitpose} &  \multicolumn{1}{c|}{-} & ViT-B & 90.0M &17.9& 256$\times$192 & 75.7 & 81.0 \\ 	

ProbPose$\dagger$~\cite{Probpose} &
-&ProbPose-S&40.3M&6.4&256$\times$192&66.1&69.7\\

			SimpleBaseline~\cite{SimpleBaseline} &  \multicolumn{1}{c|}{-}& ResNet-50 & 34.0M &5.5& 256$\times$192 & 71.8 & 77.4 \\ 
MSPN~\cite{MSPN} &  \multicolumn{1}{c|}{-}& mspn\_50  & 25.1M &5.1& 256$\times$192 & 72.3& 78.8\\  
			RSN~\cite{RSN} &  \multicolumn{1}{c|}{-}& RSN-50  & 19.3M&4.1 & 256$\times$192 & 72.4&79.0\\ 
			\hline
shufflenetv2~\cite{ShufflenetV2} &  \multicolumn{1}{c|}{-}& shufflenetv2  & 7.6M&1.4 & 256$\times$192 & 60.2&66.8\\ 
			GTPT~\cite{GTPT} & -&GTPT-S&5.4M&1.6&256$\times$192&73.6&78.9 \\
			\hline

			\multicolumn{7}{c}{Baselines with RHFR} \\  			\hline
		
		\multirow{1}{*}{Hourglass} & $G=[2,2]$ & Hourglass-52 & 95.5(+0.7)M&31.2(+2.5) & 256$\times$256 &72.9 (\textcolor{red}{$\uparrow$0.3})&78.2 \\ 
		\hline

		SWIN~\cite{SWIN} & $G=[2,2]$& swin\_b  & 93.2(+0.2)M &19.5(+0.5)& 256$\times$192 &74.0 (\textcolor{red}{$\uparrow$0.3}) &79.6\\ 
		\hline
		
		\multirow{1}{*}{ViTPose}
		& $G=[2,2]$ & ViT-B & 90.6(+0.6)M &19.8(+1.9)& 256$\times$192 & 75.9 (\textcolor{red}{$\uparrow$0.2})&81.0 \\ \hline
				ProbPose~\cite{Probpose} & $G=[2,2]$&ProbPose-S& 41.7(+1.4)M &6.7(+0.3)& 256$\times$192 &  67.7 (\textcolor{red}{$\uparrow$1.6})&71.1\\ 

\hline

			\multirow{1}{*}{SimpleBaseline}
			& $G=[2,2]$ & ResNet-50 & 34.04(+0.04)M &5.9(+0.4)& 256$\times$192 & 72.2 (\textcolor{red}{$\uparrow$0.4})&77.7\\

			\hline
		
			MSPN~\cite{MSPN} & $G=[2,2]$& MSPN\_50  & 25.7(+0.6)M &7.0(+1.9)& 256$\times$192 &73.0 (\textcolor{red}{$\uparrow$0.7}) &79.4\\ 
\hline
RSN~\cite{RSN} & $G=[2,2]$& RSN\_50  & 19.4(+0.1)M &4.2(+0.1)& 256$\times$192 & 72.8 (\textcolor{red}{$\uparrow$0.4}) &79.4\\ 
\hline

			shufflenetv2~\cite{ShufflenetV2} & $G=[2,2]$& shufflenetv2  & 7.7(+0.1)M &1.5(+0.1)& 256$\times$192 & 61.6 (\textcolor{red}{$\uparrow$1.4}) &68.1\\ 
			\hline
			GTPT~\cite{GTPT} & $G=[2,2]$&GTPT-S& 5.5(+0.1)M &2.1(+0.5)& 256$\times$192 & 73.8 (\textcolor{red}{$\uparrow$0.2})  &79.0\\ 
						\hline

		\end{tabular}
\end{table*}

\begin{figure}[t]
	\centering
	\subfloat[]{
		\begin{tikzpicture}[scale=0.42]			
			\begin{axis}[
				xlabel={Depth},
				xlabel style={font=\small},
				ylabel={Params (M)},
				ylabel style={font=\small,anchor=east,at={(0.05,0.65)}},
				xtick={1,2,3,4,5,6,7},
				axis y line*=left,
				axis x line*=bottom,
				ymin=0.5,
				ymax=1,
				legend style={
					at={(0.02,0.98)},
					anchor=north west,
					draw=none,
					fill=none
				}
				]
				\addplot[
				myteal,
				mark=triangle,
				thick
				] coordinates{
					(1,0.59238)
					(2,0.62541)
					(3,0.64205)
					(4,0.65050 )
					(5,0.65485 )
					(6,0.65715 )
					(7,0.65843)
				};
				\addlegendentry{Params}				
			\end{axis}
			\begin{axis}[
				ylabel={GFLOPs},
				ylabel style={font=\small,anchor=west,at={(1.3,0.4)}},
				axis y line*=right,
				axis x line*=top,
				xtick=\empty,
				ymin=1.8,
				ymax=2.1,
				legend style={
					at={(0.02,0.90)},
					anchor=north west,
					draw=none,
					fill=none
				}
				]
				
				\addplot[
				myorange,
				mark=*,
				thick
				] coordinates{
					(1,1.82295)
					(2,1.92676)
					(3,1.98024)
					(4,2.00855)
					(5,2.02428)
					(6,2.03371)
					(7,2.04000)
				};
				\addlegendentry{GFLOPs}
				
			\end{axis}
		\end{tikzpicture}
		\label{figz1}
	}
		\hspace{2pt}
	\subfloat[]{
		\begin{tikzpicture}[scale=0.42]
			
			\begin{axis}[
				xlabel={The numbers of nodes},
				xlabel style={font=\small},
				ylabel={Params (M)},
				ylabel style={font=\small,anchor=east,at={(0.05,0.65)}},
				xtick={1,2,3,4,5,6,7},
				xticklabels={{7},{13},{25},{49},{97},{193},{385}},
				axis y line*=left,
				axis x line*=bottom,
				ymin=0.5,
				ymax=1,
				legend style={
					at={(0.02,0.98)},
					anchor=north west,
					draw=none,
					fill=none
				}
				]
				
				\addplot[
				myteal,
				mark=triangle,
				thick
				] coordinates{
					(1,0.62541)
					(2,0.60902)
					(3,0.60083)
					(4,0.59674)
					(5,0.59469)
					(6,0.59366)
					(7,0.59315)
				};
				\addlegendentry{Params}
				
			\end{axis}
			
			\begin{axis}[
				ylabel={GFLOPs},
				ylabel style={font=\small,anchor=west,at={(1.3,0.4)}},
				axis y line*=right,
				axis x line*=top,
				xtick=\empty,
				ymin=1.8,
				ymax=2.1,
				legend style={
					at={(0.02,0.90)},
					anchor=north west,
					draw=none,
					fill=none
				}
				]
				
				\addplot[
				myorange,
				mark=*,
				thick
				] coordinates{
					(1,1.92676)
					(2,1.87957)
					(3,1.86070)
					(4,1.86070)
					(5,1.87957)
					(6,1.92676)
					(7,2.02585)
				};
				\addlegendentry{GFLOPs}
				
			\end{axis}
			
		\end{tikzpicture}
		\label{figz2}
	}
\caption{Parameter count and computational complexity of RHFR under different configurations, with input feature size $L\times C$ ($L=64\times48$, $C=256$). (a) Depth: $G=[g_d,\dots,g_1]$, where $g_i=2$ and $d=1,\dots,7$. (b) Breadth: $G=[2^n,2]$, where $n=1,\dots,7$.}
	\label{figz}
\end{figure}

 \begin{table*}[!t]
    \caption{Comparisons on the COCO validation set. The best, second-best, and third-best results are highlighted in bold, underlined, and italic, respectively. $\dagger$ means the results of our reproduction.}
    \label{tb2}
    \centering
    \begingroup
    \small
    \setlength{\tabcolsep}{5pt}
    \begin{tabular}{l|l|c|c|c|c}
        \hline
        Method & Backbone & \#Params & Input size & MAP & MAR \\ \hline
                SP-HRNet~\cite{SPHRNet2025} & SP-HRNet-256 & 13.1M & 256$\times$192 & 73.7 & 77.6 \\
                        SPTPose-B~\cite{SPTPose2026} & SPTPose-B & 13.2M & 256$\times$192 & 74.8 & 80.1 \\
        ViTPose-B~\cite{Vitpose} & ViT-B & 86.0M & 256$\times$192 & 75.8 & 81.1 \\
      SimCC~\cite{Simcc} & HRNet-W48 & 66.3M & 256$\times$192 & 75.9 & 81.2\\
        TokenPose-L/D24~\cite{TokenPose} & TokenPose-L/D24 & 27.5M & 256$\times$192 & 75.8 & 80.9 \\
        EMpose~\cite{EMpose} & HRNet-W32 & 30.3M & 256$\times$192 & 75.0 & 80.2 \\
 


        Dyn-M2-H48~\cite{DynPose2025} & Dyn-M2-H48  & -- & 256$\times$192 & 75.0 & 80.3 \\
                HRNet~\cite{Hrnet} & HRNet-W32 & 28.5M & 256$\times$192 & 74.4 & 79.8 \\
             PGBS~\cite{PGBS} & HRNet-W32 & 81.0M & 256$\times$192 & 75.6 & 80.6 \\
        \hline
        Ours-small & HRNet-W32 & 37.1M & 256$\times$192 & 75.1 & 80.1 \\
        Ours-large & HRNet-W32 & 50.7M & 256$\times$192 & 75.8 & 80.8 \\
        \hline
        SimpleBaseline~\cite{SimpleBaseline} & ResNet-152 & 68.6M & 384$\times$288 & 74.3 & 79.7 \\
        ViTPose-B$\dagger$~\cite{Vitpose} & ViT-B & 86.0M & 384$\times$288 & 76.8& 81.8 \\
  
        HRNet~\cite{Hrnet} & HRNet-W48 & 63.6M & 384$\times$288 & 76.3 & 81.2 \\
        DiffusionPose~\cite{DiffusionPose} & HRNet-W48 & 74.0M & 384$\times$288 & \bf 77.6 & 82.4 \\
        SHaRPose~\cite{SHaRPose} & SHaRPose-Base & 118.1M & 384$\times$288 & \underline {77.4} & 82.4 \\


        DE-HRNet ~\cite{DEHRNet2025} & HRNet-W48 & 74.8M & 384$\times$288 & 76.6 & 81.6 \\
                HRNet~\cite{Hrnet} & HRNet-W32 & 28.5M & 384$\times$288 & 75.8 & 81.0 \\
        PGBS~\cite{PGBS} & HRNet-W32 & 81.0M & 384$\times$288 & 76.9 & 81.7 \\

        \hline
        Ours-small & HRNet-W32 & 37.1M & 384$\times$288 & 76.7 & 81.5 \\
        Ours-large & HRNet-W32 & 50.7M & 384$\times$288 & \it{77.0} & 81.7 \\
        \hline
    \end{tabular}
    \endgroup
\end{table*}

 \begin{table}[!t]	
 	\caption{Comparisons on CrowdPose test set with YOLOv3~\cite{Yolov3} human detector. * denotes using a stronger Faster RCNN~\cite{FastRCNN} detector. $\dagger$ denotes  the results of our reimplementation. Bold indicates the best result.}
 	\label{tb3}
 	\centering
 	\begin{tabular}{c|c|c|c}
 		\hline
 		Method     & Backbone & Input size & MAP \\ \hline
 		Sim.Base.~\cite{SimpleBaseline}& ResNet-152 & 256$\times$192     & 65.6  \\
 		ViTPose$\dagger$~\cite{Vitpose}& ViT-B & 256$\times$192     & 66.3  \\
 		LJOF~\cite{LJOF}&	ViT-B& 256$\times$192&67.1\\
 		PGBS~\cite{PGBS} & HRNet-W32  & 256$\times$192 & 68.9      \\
 			 		HRNet~\cite{Hrnet}& HRNet-W32 & 256$\times$192     & 67.5  \\
 		ViTPose$\dagger$~\cite{Vitpose}& ViT-B & 384$\times$288    & 68.6  \\
 		MIPNet~\cite{Mipnet}&ResNet-101&384$\times$288&68.1\\
 		MIPNet*~\cite{Mipnet}&ResNet-101&384$\times$288&70.0\\
 		HRNet*~\cite{Hrnet}& HRNet-W48 & 384$\times$288     & 69.3  \\ 
 			PoseNet++~\cite{PoseNet++}& SHN& 384$\times$288     & 70.6  \\ 
 	
 		PGBS~\cite{PGBS}& HRNet-W32  & 384$\times$288 & 70.5   \\
 		\hline
 		Ours-small & HRNet-W32 & 256$\times$192 & 68.4 \\
 		Ours-large & HRNet-W32  & 256$\times$192 & 69.0\\
 		Ours-small & HRNet-W32  & 384$\times$288  &70.0 \\
 		Ours-large & HRNet-W32  & 384$\times$288 &{\bf{70.6}}\\ \hline
 		
 	\end{tabular}
 	
 \end{table}

 \begin{table*}[!t]	
 	\caption{Comparisons of PCKh@0.5 scores on the MPII test set. $\dagger$ denotes our replicated results. {Bold} indicates the best result.}
 	\label{tb4}
 	\centering
\begin{tabular}{c|ccccccc|c}
 			\hline
 			Method  & Head & Sho. & Elb. & Wri. & Hip  & Knee & Ank. & Mean \\ \hline
			Xu et al.~\cite{HRPVT}  & 96.9& 96.1   & 90.3  & 84.9  & 89.7& 86.2 & 81.7& 89.9 \\
 			Luvizon et al.~\cite{PartDetectionAndContextualInformation} & 98.1 & 96.6    & 92.0  & 87.5  & 90.6 & 88.0 & 82.7 & 91.2 \\ 
 	
 			Wang et al.~\cite{AttentionRefinedNetwork} & 98.3 & 96.7    & 92.4  & 88.5  & 90.4 & 88.3 & 84.4 & 91.6 \\ 
 			Chou et al.~\cite{SelfAdversarialTraining} & 98.2 & 96.8    & 92.2  & 88.0  & 91.3 & 89.1 & 84.9 & 91.8 \\ 
 			Chen et al.~\cite{AdversarialLearningOfStructureAware} & 98.1& 96.5   & 92.5  & 88.5  & 90.2 & 89.6 & 86.0 & 91.9 \\ 
 			Tang et al.~\cite{DeeplyLearnedCompositionalModels}& 98.4 & 96.9    & 92.6  & 88.7 &\bf 91.8 & 89.4 & 86.2  &  92.3 \\ 
 			Zhang et al.~\cite{MSSA_ADA}& 98.5 & 96.8    & 93.0 & 88.7 & 91.2& 89.3 & 85.4  &  92.2 \\ 
 			Lv et al.~\cite{PoseNet++}& 98.6 & \bf 97.0     & 92.9  & 88.6  & 91.6 & 89.5& \bf 86.3&  92.4 \\ 
			TokenPose-L/D6$\dagger$~\cite{TokenPose}    & 98.4 & 96.3   & 91.7& 87.2 & 90.5& 87.7 & 83.5& 91.1\\ 	
 			HRNet-W32$\dagger$& 97.9 & 96.5     & 92.3  & 88.2  & 90.9 & 88.1 & 83.8  &  91.5 \\ 
 			PGBS~\cite{PGBS}     & 98.5  & 96.7  & 92.8& 88.6    & 91.1 & 89.2 & 85.2  & 92.1\\		\hline
 			Ours-small     & \bf 98.6  &  96.9   & \bf 93.1& \bf 89.3  & 91.6 &\bf 89.9& 85.9  & \bf 92.5
 			\\	\hline
 	\end{tabular}
 	
 \end{table*}

 \begin{figure}[!t]
 	\centering
 	\subfloat[]{\includegraphics[width=0.6in]{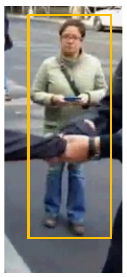}%
 		\label{fig0-a}}
 	\hspace{0.3cm}
 	\subfloat[]{\includegraphics[width=1.8in]{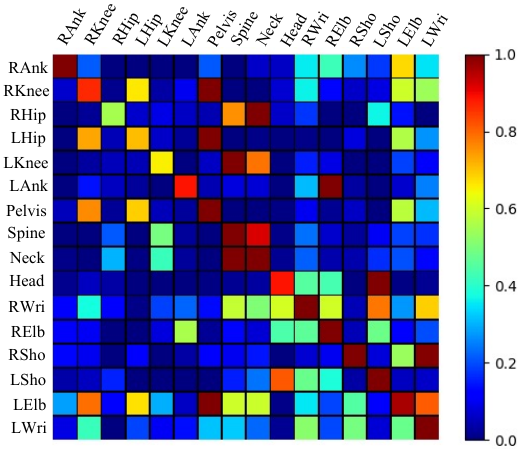}%
 		\label{fig0-b}}
 	
\caption{(a) Input image from the MPII validation set. (b) Visualization of context relations among joints in (a). Each row shows the relations of one joint with all other joints, where higher scores indicate stronger relations.}
 	\label{fig0}
 \end{figure}
 
 \begin{figure*}[h]
 	\centering
 	
 	\includegraphics[width=0.8\linewidth]{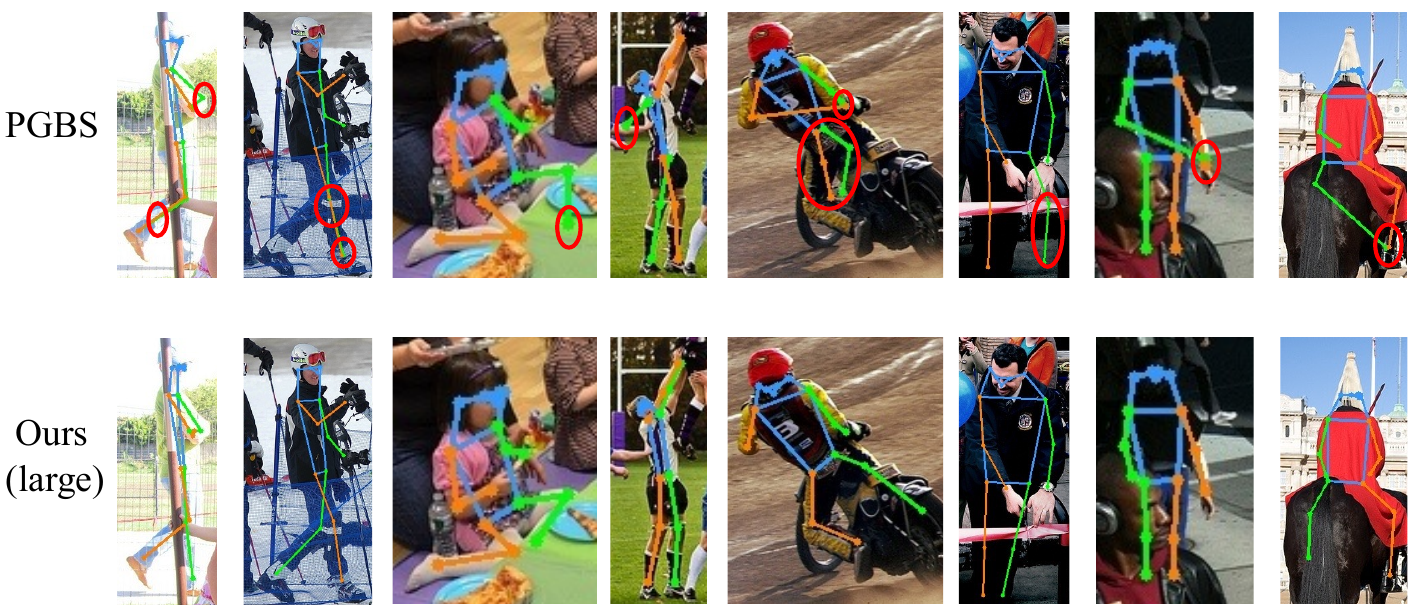}
 	
 	\caption{Comparison of the visualization results between our large network and the PGBS~\cite{PGBS} on the COCO val set. The input images are all 384$\times$288.}
 	\label{fig-1}
 \end{figure*}

 \begin{figure*}[h]
 	\centering
 	
 	\includegraphics[width=1\linewidth]{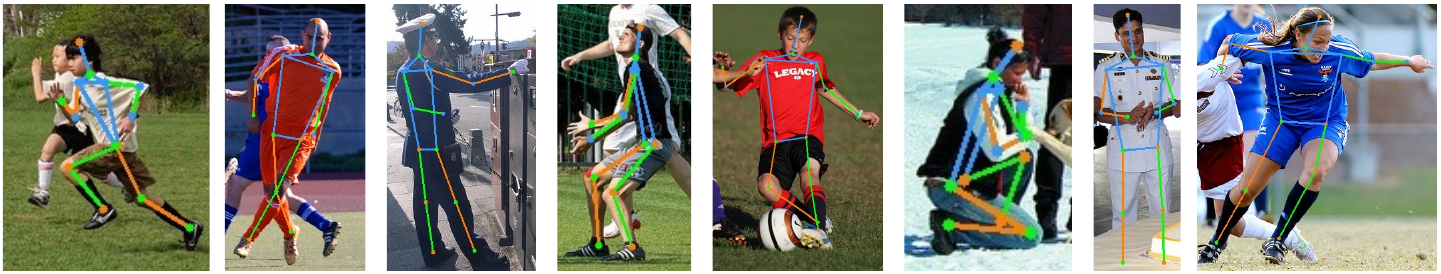}
 	
 	\caption{The visualization results of our large network on the CrowdPose test set. The input images are 384$\times$288.}
 	\label{fig:more_vis}
 \end{figure*}

 \begin{figure}[h]
 	\centering
 	
 	\includegraphics[width=0.6\linewidth]{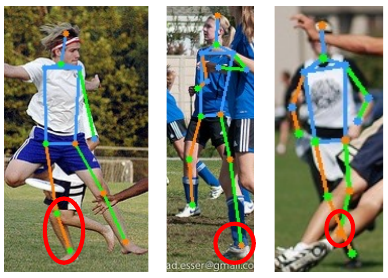}
 	
\caption{Failure cases of our large network under severe occlusion and joint interference. Input images are resized to $384\times288$.}
 	\label{fig:fail_example}
 \end{figure}
 
%
%

\begin{table}
\caption{Ablation of different RHFR configurations in the plug-and-play setting on COCO val set.}
	\label{tb:AblationofRMPG}
	\centering
	\begin{tabular}{c|l|c}
		\hline
		Method & RHFR   & MAP  \\ 
		\hline
		
		\multirow{6}{*}{SimBa.~\cite{SimpleBaseline}} 
		
		& ${G}=[2]$  & 71.6\\
		& ${G}=[2,2]$ & 72.2\\
		& ${G}=[2,2,2]$&72.0\\
\cline{2-3}
		& ${G}=[4]$& 71.7\\
		& ${G}=[4,4]$& 71.8\\
		& ${G}=[4,2]$& 71.7\\
				& ${G}=[4,4,4]$& 71.8\\
		\hline
	\end{tabular}
\end{table}
 
 \begin{table}[!t]	
 	\caption{Ablation experiments on the MPII test set analyze the impact under different configurations of RHFR$_\text{s}$ and RHFR$_\text{u}$. $\emptyset$ denotes removing supervision in RHFR$_\text{s}$, $\nabla$ disables NA and * means removing body-level supervision on $F_0$.}
 	\label{tb5}
 	\centering
 	\begin{tabular}{c|ll|c}
 		\hline
 		Method  & RHFR$_\text{s}$& RHFR$_\text{u}$&Mean \\ \hline
 		\multirow{11}{*}{Ours network}&  \multirow{7}{*}{$G=[5,2]$}  & $ G=[1]$   & 91.7\\	 
 		&  & $ G=[2]$& 92.0\\	 
 		& & $ G=[4]$  & 91.9\\
 		
 		&& $ G=[4,4]$ & 91.9\\
 		& &$ G=[2,2]$ &{\bf{92.2}}\\
 		& & $ G=[2,2,2] $ &91.9 \\
 		& & \multicolumn{1}{c|}{-}&91.8 \\  \cline{2-4}
 		&$ G=[5,8]$ &  $ G=[2,2]$&91.9\\
 		&$ G=[5,4]$ &  $ G=[2,2]$&92.0\\
 		& $ G=[5,2]_{\emptyset}$&  $ G=[2,2]$&91.6\\
 		&$ G=[5,2]_\nabla$&$ G=[2,2]_\nabla$ &91.4\\ \hline
 		 Ours network*&  $G=[5,2]_{\emptyset}$  & $ G=[2,2]$   &91.5 \\ \hline	 
 		HRNet-W32$\dagger$  &\multicolumn{2}{c|}{--}    & 91.2\\ 	
 		\hline
 	\end{tabular}
 	
 \end{table}

\section{Experiments}

\subsection{Datasets and evaluation methods}
\label{sec:exp}


{\bf{Datasets.}} For the CrowdPose datasets, there are $20$k images and $80$k human instances labeled with $14$ joints and the training, validation and testing subset are split in proportional to 5:1:4~\cite{Crowdpose}. For the COCO keypoint detection dataset, there are more than $200$k images and $250$k person instances, labeled with $17$ joints, of which $57$k images are used for training, $5$k images are used for validation. For the MPII Human Pose dataset, there are approximately $25$k images and $40$k annotated samples with $16$ joints per instance, of which $28$k are used for training and $11$k for testing.

{\bf{Evaluation methods.}}~For CrowdPose and COCO datasets, we use mean average precision (MAP) and mean average recall (MAR) when evaluating the model. In contrast, the MPII dataset uses PCKh score to evaluate the accuracy of pose estimation.

\subsection{Implementation details}
\label{Sec.4.2}
{\bf{Training and testing of our network.}} For the CrowdPose and COCO datasets, all input images are resized to either $256\times 192$ or $384\times 288$ resolution. During verification and testing, we employ the YOLOv3 human detector~\cite{Yolov3} for CrowdPose and utilize detected person boxes from SimpleBaseline~\cite{SimpleBaseline} for COCO. In the case of the MPII dataset, images are resized to $256\times 256$ resolution, and testing involves using provided person boxes along with a six-scale pyramid testing method~\cite{LearningFeaturePyramids}. Other training and testing strategies are consistent with HRNet~\cite{Hrnet}.

In our hierarchical network, for supervised RHFR$\text{s}$, setting $G=[5,2]$. For unsupervised RHFR$\text{u}$, setting $G=[2,2]$.

\subsection{RHFR performance}
As shown in Table \ref{tb1}, integrating RHFR consistently improves the performance of various backbone networks. Shufflenetv2~\cite{ShufflenetV2} achieves 1.4 MAP improvement, while ProbPose with ProbPose-S backbone achieves 1.6 MAP improvement, MSPN with backbone MSPN\_50 achieves 0.7 MAP improvement. For other method, such as ViTPose, Hourglass, and RSN, also show corresponding improvements.

To investigate the underlying mechanism behind the performance improvement brought by RHFR, we visualize the feature response changes during successive refinement by RHFR based on the SimpleBaseline experiments in Table~\ref{tb1}, as shown in Fig.~\ref{figrhfr}. 
Specifically, the feature maps are averaged along the channel dimension and then overlaid onto the corresponding input image to visualize their spatial response distributions.
The feature responses become progressively more concentrated around the human body as the hierarchical refinement proceeds. Compared with the input feature in Fig.~\ref{figrhfr-b}, the refined parent node features in Fig.~\ref{figrhfr-c} and Fig.~\ref{figrhfr-d} exhibit increasingly focused responses to human-related regions. This trend is further strengthened in the refined root node feature in Fig.~\ref{figrhfr-e}, where the responses are more concentrated on the overall human body. These observations suggest that the hierarchical reasoning in RHFR progressively enhances human-related representations while suppressing irrelevant background responses, providing more discriminative features for pose estimation.

{\bf{Parameter analysis.}} 
As shown in Table~\ref{tb1}, the additional parameters introduced by RHFR generally remain below 1M across different backbone architectures. In particular, RHFR introduces only 0.04M additional parameters to SimpleBaseline, demonstrating its low parameter overhead. In contrast, the computational overhead is closely related to the spatial resolution of the input feature map. 
For example, the RHFR module in Hourglass takes a $64\times64$ feature map as input, the largest spatial resolution among all evaluated models, resulting in an additional computational cost of 2.5 GFLOPs. 
These results demonstrate that RHFR achieves feature refinement with a relatively small parameter overhead, while its computational cost varies with the input feature characteristics.

{\bf Effect of hierarchical depth and breadth on complexity.}
Fig.~\ref{figz1} and~\ref{figz2} further analyze the impact of hierarchy depth and the number of nodes. With increasing hierarchy depth, both the number of parameters and GFLOPs increase, eventually stabilizing.
In contrast, increasing the number of nodes leads (breadth) to a reduction in the number of parameters, while the GFLOPs first decrease and then gradually increase. This suggests that a moderate increase in breadth can improve parameter and computation efficiency, whereas excessively large breadth introduces additional computational overhead.
Overall, RHFR exhibits favorable scalability with respect to both depth and breadth.

\subsection{Hierarchical modeling network performance}
Our network is evaluated in two variants, small and large. The variants differ in the number of convolutional layers.

{\bf{COCO keypoint detection benchmark.}} Table~\ref{tb2} shows the results of our network and existing methods on the COCO validation set. Our {\bf small network} achieves 75.1 AP at 256$\times$192 resolution and 76.7 AP at 384$\times$288 resolution, outperforming HRNet-W32 by 0.7 and 0.9 AP, respectively. Our {\bf large network} achieves 75.8 AP and 77.0 AP at the two resolutions, respectively, and outperforms PGBS with fewer parameters. It also achieves a higher AP than ViTPose-B with fewer parameters. Although DiffusionPose and SHaRPose achieve slightly higher APs of 77.6 and 77.4, respectively, they require 74.0M and 118.1M parameters, respectively, whereas our {\bf large network} achieves 77.0 AP with only 50.7M parameters, demonstrating a more favorable trade-off between accuracy and parameter count.

{\bf{CrowdPose benchmark.}} Table~\ref{tb3} shows the results on the CrowdPose test set. Our {\bf small network} achieves 68.4 MAP at 256$\times$192 resolution and 70.0 MAP at 384$\times$288 resolution, outperforming HRNet-W32 by 0.9 and HRNet-W48 by 0.7 respectively, and outperforming ViT-B by 2.1 and 1.4 respectively. 
The {\bf large network} achieves 69.0 AP and 70.6 AP at the two input resolutions, respectively, outperforming PGBS with fewer parameters. These results demonstrate that RHFR enables effective explicitly human structure modeling with fewer parameters, and improves pose estimation performance in CrowdPose datesets, which features crowded scenes.

{\bf{MPII benchmark.}} As shown in Table~\ref{tb4}, our {\bf small network} achieves 92.5 PCKh@0.5 with 256$\times$256 inputs, outperforming HRNet-W32 and PGBS by 1.0 and 0.4 points, respectively. Notably, our method achieves the best results on the elbow, wrist, and knee joints, while maintaining competitive performance on the ankle joints. These results demonstrate that RHFR not only improves overall pose estimation accuracy but also enhances the prediction of challenging joints.

\textbf{Visualization.} To further investigate whether RHFR can explicitly capture structural context in our hierarchical network, we take joint-level context relations as an example and visualize the correlations among different joints in Fig.~\ref{fig0-b}. Specifically, based on the second RHFR$_\text{s}$, we extract feature vectors at the ground-truth locations of individual joints, compute pairwise similarities via inner products, and normalize the resulting similarity matrix by the maximum value of each row. As shown in Fig.~\ref{fig0-b}, the matrix exhibits strong self-correlations along the diagonal. Meanwhile, occluded joints also show strong correlations with other joints, indicating that the model can leverage context information from related joints to complement their feature representations. These results demonstrate that supervising RHFR$_\text{s}$ with joint-level supervision enables it to effectively capture context relations among human joints, thereby improving pose estimation in challenging scenarios. As shown in Fig.~\ref{fig-1} and Fig.~\ref{fig:more_vis}, our large network achieves more accurate predictions under severe occlusion and in crowded scenes. 
However, as shown in Fig. \ref{fig:fail_example}, failures still occur under severe occlusion, particularly when target joints are almost completely invisible and interfered with by overlapping body parts from other individuals. This is because hierarchical feature refinement in RHFR relies on available visual cues, which become insufficient to distinguish target joints under such challenging conditions.

\subsection{Ablation study}

\subsubsection{Ablation of the RHFR in the plug-and-play experiments}
To determine the optimal RHFR configuration for the plug-and-play experiments in Table~\ref{tb1}, we conduct ablation experiments using SimpleBaseline (ResNet-50) as the backbone. As shown in Table~\ref{tb:AblationofRMPG}, ${G}=[2,2]$ achieves the best performance of 72.2 mAP, outperforming all other configurations. These results demonstrate that ${G}=[2,2]$ provides the most effective hierarchical configuration and is therefore adopted in Table~\ref{tb1}.

\subsubsection{Ablation of the Hierarchical Network with RHFR}

Our ablation experiments are conducted on the MPII test set without multi-scale testing~\cite{LearningFeaturePyramids}, using an input size of $256\times256$, as shown in Table~\ref{tb5}.

\textbf{The influence of $G$.} For RHFR$_{\text{u}}$, $G=[2,2]$ achieves the best performance of 92.2, outperforming other configurations. For RHFR$_{\text{s}}$, we fix the first-level branching factor to 5 to maintain consistency with PGBS, which partitions the human body into five semantic regions. Increasing the second-level branching factor from 2 to 4 and 8 reduces the performance from 92.2 to 92.0 and 91.9, respectively, suggesting that excessive decomposition may introduce redundant contextual interactions. Thus, we adopt $G=[5,2]$ for RHFR$_{\text{s}}$ and $G=[2,2]$ for RHFR$_{\text{u}}$.

\textbf{The effectiveness of RHFR$_{\text{u}}$.} Removing RHFR$_{\text{u}}$ reduces the performance from 92.2 to 91.8, demonstrating the benefit of unsupervised hierarchical refinement.

\textbf{The effectiveness of supervision.} Removing limb and joint supervision from RHFR$_{\text{s}}$ decreases the performance from 92.2 to 91.6. Further removing the body-level supervision on $F_0$ results in a score of 91.5, highlighting the importance of multi-level structural supervision.

\textbf{The effectiveness of NA.} Disabling context relations in both RHFR$_{\text{s}}$ and RHFR$_{\text{u}}$ reduces the performance to 91.4, demonstrating the effectiveness of NA.

\section{Conclusion}
\label{sec:conc}
The RHFR module provides new methods for feature map optimization while efficiently helping to explicit model context relations and hierarchies among body parts. We hope that the RHFR module can be widely used in various tasks. Future work could explore the implementation of RHFR-based backbone networks, or better decomposition and composition strategies.

\bibliographystyle{IEEEtran}
\bibliography{reference.bib} 

@String(IJCV = {Int. J. Comput. Vis.})

@String(CVPR= {IEEE Conf. Comput. Vis. Pattern Recog.})

@String(ICCV= {Int. Conf. Comput. Vis.})

@String(ECCV= {Eur. Conf. Comput. Vis.})

@String(NIPS= {Adv. Neural Inform. Process. Syst.})

@String(TMM  = {IEEE Trans. Multimedia})

@String(AAAI = {AAAI})

@String(TPAMI  = {IEEE TPAMI})

@String(IJCV  = {IJCV})

@String(CVPR  = {CVPR})

@String(ICCV  = {ICCV})

@String(ECCV  = {ECCV})

@String(NIPS  = {NeurIPS})

@String(TCSVT = {IEEE TCSVT})

@String(TMM   =	{IEEE TMM})

@string{CVPR = "{Proc. IEEE/CVF Conf. Comput. Vis. Pattern Recognit. (CVPR)}"}

@string{ECCV = "{Proc. Eur. Conf. Comput. Vis. (ECCV)}"}

@string(IF = {Information Fusion})

@string(ACML = {ACML})

@string(APSIPA_ASC = {APSIPA ASC})

@string(FTCGV = {Found. Trends Comput. Graph. Vision (FTCGV)})

@string(NEUROCOMP = {Neurocomputing})

@string(COMP_GRAPH = {Comput. Graph.})

@string(MultimediaSystems = {Multimedia Syst.})

@string(NeuralProcessingLetters = {Neural Process. Lett.})

@string(NeuralComputingandApplications = {Neural Comput. Appl.})

@string{DAS = {Discov. Appl. Sci.}}

@inproceedings{SimpleBaseline,
	title={Simple baselines for human pose estimation and tracking},
	author={Xiao, Bin and Wu, Haiping and Wei, Yichen},
	booktitle=ECCV,
	pages={466--481},
	year={2018}
}

@inproceedings{Hrnet,
	title={Deep high-resolution representation learning for human pose estimation},
	author={Sun, Ke and Xiao, Bin and Liu, Dong and Wang, Jingdong},
	booktitle=CVPR,
	pages={5693--5703},
	year={2019}
}

@inproceedings{Mipnet,
	title={Multi-instance pose networks: Rethinking top-down pose estimation},
	author={Khirodkar, Rawal and Chari, Visesh and Agrawal, Amit and Tyagi, Ambrish},
	booktitle=ICCV,
	pages={3122--3131},
	year={2021}
}

@inproceedings{DeepFullyConnected,
	title={Deep fully-connected part-based models for human pose estimation},
	author={De Bem, Rodrigo and Arnab, Anurag and Golodetz, Stuart and Sapienza, Michael and Torr, Philip},
	booktitle=ACML,
	pages={327--342},
	year={2018},
	organization={PMLR}
}

@article{GrammarOfImages,
	title={A stochastic grammar of images},
	author={Zhu, Song-Chun and Mumford, David and others},
	journal=FTCGV,
	volume={2},
	number={4},
	pages={259--362},
	year={2007},
	publisher={Now Publishers, Inc.}
}

@inproceedings{DeeplyLearnedCompositionalModels,
	title={Deeply learned compositional models for human pose estimation},
	author={Tang, Wei and Yu, Pei and Wu, Ying},
	booktitle=ECCV,
	pages={190--206},
	year={2018}
}

@inproceedings{Hourglass,
	title={Stacked hourglass networks for human pose estimation},
	author={Newell, Alejandro and Yang, Kaiyu and Deng, Jia},
	booktitle=ECCV,
	pages={483--499},
	year={2016},
	organization={Springer}
}

@article{AdversarialLearningOfStructureAware,
	title={Adversarial learning of structure-aware fully convolutional networks for landmark localization},
	author={Chen, Yu and Shen, Chunhua and Chen, Hao and Wei, Xiu-Shen and Liu, Lingqiao and Yang, Jian},
	journal=TPAMI,
	volume={42},
	number={7},
	pages={1654--1669},
	year={2019},
	publisher={IEEE}
}

@inproceedings{MultiScaleStructureAware,
	title={Multi-scale structure-aware network for human pose estimation},
	author={Ke, Lipeng and Chang, Ming-Ching and Qi, Honggang and Lyu, Siwei},
	booktitle=ECCV,
	pages={713--728},
	year={2018}
}

@inproceedings{TokenPose,
	title={Tokenpose: Learning keypoint tokens for human pose estimation},
	author={Li, Yanjie and Zhang, Shoukui and Wang, Zhicheng and Yang, Sen and Yang, Wankou and Xia, Shu-Tao and Zhou, Erjin},
	booktitle=ICCV,
	pages={11313--11322},
	year={2021}
}

@inproceedings{RSN,
	title={Learning delicate local representations for multi-person pose estimation},
	author={Cai, Yuanhao and Wang, Zhicheng and Luo, Zhengxiong and Yin, Binyi and Du, Angang and Wang, Haoqian and Zhang, Xiangyu and Zhou, Xinyu and Zhou, Erjin and Sun, Jian},
	booktitle=ECCV,
	pages={455--472},
	year={2020},
	organization={Springer}
}

@inproceedings{LearningFeaturePyramids,
	title={Learning feature pyramids for human pose estimation},
	author={Yang, Wei and Li, Shuang and Ouyang, Wanli and Li, Hongsheng and Wang, Xiaogang},
	booktitle=ICCV,
	pages={1281--1290},
	year={2017}
}

@article{KnowledgeGuide,
	title={Knowledge-guided deep fractal neural networks for human pose estimation},
	author={Ning, Guanghan and Zhang, Zhi and He, Zhiquan},
	journal=TMM,
	volume={20},
	number={5},
	pages={1246--1259},
	year={2017},
	publisher={IEEE}
}

@article{PartDetectionAndContextualInformation,
	title={Human pose regression by combining indirect part detection and contextual information},
	author={Luvizon, Diogo C and Tabia, Hedi and Picard, David},
	journal=COMP_GRAPH,
	volume={85},
	pages={15--22},
	year={2019},
	publisher={Elsevier}
}

@inproceedings{SelfAdversarialTraining,
	title={Self adversarial training for human pose estimation},
	author={Chou, Chia-Jung and Chien, Jui-Ting and Chen, Hwann-Tzong},
	booktitle=APSIPA_ASC,
	pages={17--30},
	year={2018},
	organization={IEEE}
}

@article{EMpose,
	title={Body parts relevance learning via expectation--maximization for human pose estimation},
	author={Yue, Luhui and Li, Junxia and Liu, Qingshan},
	journal=MultimediaSystems,
	volume={27},
	number={5},
	pages={927--939},
	year={2021},
	publisher={Springer}
}

@article{AttentionRefinedNetwork,
	title={Attention refined network for human pose estimation},
	author={Wang, Xiangyang and Tong, Jiangwei and Wang, Rui},
	journal=NeuralProcessingLetters,
	volume={53},
	number={4},
	pages={2853--2872},
	year={2021},
	publisher={Springer}
}

@article{SpatialContextual,
	title={Human pose estimation with spatial contextual information},
	author={Zhang, Hong and Ouyang, Hao and Liu, Shu and Qi, Xiaojuan and Shen, Xiaoyong and Yang, Ruigang and Jia, Jiaya},
	journal={arXiv:1901.01760},
	year={2019}
}

@inproceedings{Crowdpose,
	title={Crowdpose: Efficient crowded scenes pose estimation and a new benchmark},
	author={Li, Jiefeng and Wang, Can and Zhu, Hao and Mao, Yihuan and Fang, Hao-Shu and Lu, Cewu},
	booktitle=CVPR,
	pages={10863--10872},
	year={2019}
}

@inproceedings{Vitpose,
	title={ViTPose: simple vision transformer baselines for human pose estimation},
	author={Xu, Yufei and Zhang, Jing and Zhang, Qiming and Tao, Dacheng},
	booktitle=NIPs,
	pages={38571--38584},
	year={2022}
}

@article{Yolov3,
	title={Yolov3: An incremental improvement},
	author={Redmon, Joseph and Farhadi, Ali},
	journal={arXiv:1804.02767},
	year={2018}
}

@article{FastRCNN,
	title={An implementation of faster rcnn with study for region sampling},
	author={Chen, Xinlei and Gupta, Abhinav},
	journal={arXiv:1702.02138},
	year={2017}
}

@inproceedings{PCTPose,
	title={Human Pose as Compositional Tokens},
	author={Geng, Zigang and Wang, Chunyu and Wei, Yixuan and Liu, Ze and Li, Houqiang and Hu, Han},
	booktitle=CVPR,
	pages={660--671},
	year={2023}
}

@article{HumanBodyAwareFeature,
	title={Human Body-Aware Feature Extractor Using Attachable Feature Corrector for Human Pose Estimation},
	author={Kim, Ginam and Kim, Hyunsung and Kong, Kyeongbo and Song, Jou-Won and Kang, Suk-Ju},
	journal=TMM,
	year={2022},
	publisher={IEEE}
}

@article{GroupingByCenter,
	title={Grouping by center: Predicting centripetal offsets for the bottom-up human pose estimation},
	author={Jin, Lei and Wang, Xiaojuan and Nie, Xuecheng and Liu, Luoqi and Guo, Yandong and Zhao, Jian},
	journal=TMM,
	year={2022},
	publisher={IEEE}
}

@article{EHPE,
	title={EHPE: Skeleton cues-based gaussian coordinate encoding for efficient human pose estimation},
	author={Liu, Hai and Liu, Tingting and Chen, Yu and Zhang, Zhaoli and Li, You-Fu},
	journal=TMM,
	year={2022},
	publisher={IEEE}
}

@article{Hybrid,
	title={Hybrid refinement-correction heatmaps for human pose estimation},
	author={Kamel, Aouaidjia and Sheng, Bin and Li, Ping and Kim, Jinman and Feng, David Dagan},
	journal=TMM,
	volume={23},
	pages={1330--1342},
	year={2020},
	publisher={IEEE}
}

@article{BoundingBoxLSTM,
	title={Multi-person pose estimation using bounding box constraint and LSTM},
	author={Li, Miaopeng and Zhou, Zimeng and Liu, Xinguo},
	journal=TMM,
	volume={21},
	number={10},
	pages={2653--2663},
	year={2019},
	publisher={IEEE}
}

@article{PGBS,
	title={Human Pose Estimation via Parse Graph of Body Structure},
	author={Liu, Shibang and Xie, Xuemei and Shi, Guangming},
	journal=TCSVT,
	year={2024},
	publisher={IEEE}
}

@inproceedings{ShufflenetV2,
	title={Shufflenet v2: Practical guidelines for efficient cnn architecture design},
	author={Ma, Ningning and Zhang, Xiangyu and Zheng, Hai-Tao and Sun, Jian},
	booktitle=ECCV,
	pages={116--131},
	year={2018}
}

@article{MSPN,
	title={Rethinking on multi-stage networks for human pose estimation},
	author={Li, Wenbo and Wang, Zhicheng and Yin, Binyi and Peng, Qixiang and Du, Yuming and Xiao, Tianzi and Yu, Gang and Lu, Hongtao and Wei, Yichen and Sun, Jian},
	journal={arXiv preprint arXiv:1901.00148},
	year={2019}
}

@inproceedings{SWIN,
	title={Swin transformer: Hierarchical vision transformer using shifted windows},
	author={Liu, Ze and Lin, Yutong and Cao, Yue and Hu, Han and Wei, Yixuan and Zhang, Zheng and Lin, Stephen and Guo, Baining},
	booktitle=ICCV,
	pages={10012--10022},
	year={2021}
}

@inproceedings{Simcc,
	title={Simcc: A simple coordinate classification perspective for human pose estimation},
	author={Li, Yanjie and Yang, Sen and Liu, Peidong and Zhang, Shoukui and Wang, Yunxiao and Wang, Zhicheng and Yang, Wankou and Xia, Shu-Tao},
	booktitle=ECCV,
	pages={89--106},
	year={2022},
	organization={Springer}
}

@article{MSSA_ADA,
	title={Enhancement and optimisation of human pose estimation with multi-scale spatial attention and adversarial data augmentation},
	author={Zhang, Tong and Li, Qilin and Wen, Jingtao and Chen, CL Philip},
	journal=IF,
	volume={111},
	pages={102522},
	year={2024},
	publisher={Elsevier}
}

@article{LJOF,
	title={Occluded human pose estimation based on limb joint augmentation},
	author={Han, Gangtao and Song, Chunxiao and Wang, Song and Wang, Hao and Chen, Enqing and Wang, Guanghui},
	journal=NeuralComputingandApplications,
	volume={37},
	number={3},
	pages={1241--1253},
	year={2025},
	publisher={Springer}
}

@inproceedings{GTPT,
	title={GTPT: Group-based token pruning transformer for efficient human pose estimation},
	author={Wang, Haonan and Liu, Jie and Tang, Jie and Wu, Gangshan and Xu, Bo and Chou, Yanbing and Wang, Yong},
	booktitle=ECCV,
	pages={213--230},
	year={2024},
	organization={Springer}
}

@inproceedings{Graphpcnn,
	title={Graph-pcnn: Two stage human pose estimation with graph pose refinement},
	author={Wang, Jian and Long, Xiang and Gao, Yuan and Ding, Errui and Wen, Shilei},
	booktitle=ECCV,
	pages={492--508},
	year={2020},
	organization={Springer}
}

@article{DiffusionPose,
	title={Learning structure-guided diffusion model for 2d human pose estimation},
	author={Qiu, Zhongwei and Yang, Qiansheng and Wang, Jian and Wang, Xiyu and Xu, Chang and Fu, Dongmei and Yao, Kun and Han, Junyu and Ding, Errui and Wang, Jingdong},
	journal={arXiv preprint arXiv:2306.17074},
	year={2023}
}

@article{DGN,
	title={Dual graph networks for pose estimation in crowded scenes},
	author={Tu, Jun and Wu, Gangshan and Wang, Limin},
	journal=IJCV,
	volume={132},
	number={3},
	pages={633--653},
	year={2024},
	publisher={Springer}
}

@inproceedings{SHaRPose,
	title={Sharpose: Sparse high-resolution representation for human pose estimation},
	author={An, Xiaoqi and Zhao, Lin and Gong, Chen and Wang, Nannan and Wang, Di and Yang, Jian},
	booktitle=AAAI,
	volume={38},
	number={2},
	pages={691--699},
	year={2024}
}

@article{HRPVT,
	title={HRPVT: High-Resolution Pyramid Vision Transformer for medium and small-scale human pose estimation},
	author={Xu, Zhoujie and Dai, Meng and Zhang, Qing and Jiang, Xiaodi},
	journal=NEUROCOMP,
	volume={619},
	pages={129154},
	year={2025},
	publisher={Elsevier}
}

@inproceedings{Probpose,
	title={Probpose: A probabilistic approach to 2d human pose estimation},
	author={Purkrabek, Miroslav and Matas, Jiri},
	booktitle=CVPR,
	pages={27124--27133},
	year={2025}
}

@article{DEHRNet2025,
	title={DE-HRNet: Detail enhanced high-resolution network for human pose estimation},
	author={Liu, Yuxuan and Zhou, Guohui and He, Wei and Zhu, Hailong and Cui, Yanling},
	journal={Plos one},
	volume={20},
	number={9},
	pages={e0325540},
	year={2025},
	publisher={Public Library of Science, San Francisco, CA USA}
}

@article{SPTPose2026,
	title={Enhanced human pose estimation via self-distilled and token-pruned transformer},
	author={Zhang, Jundu and Deng, Zhengjie and Li, Xiyan and Yan, Sijian and Li, Zhirui and Liu, Chang and Ke, Weidong},
	journal={Scientific Reports},
	year={2026},
	publisher={Nature Publishing Group}
}

@article{SPHRNet2025,
	title={SP-HRNet: single-path neural architecture search for human pose estimation},
	author={Sun, Mengyu and Wang, Yichao and Wang, Peng and Gao, Wuqi and Li, Qunli and Li, Xiaoyan and Di, Ruohai},
	journal=DAS,
	volume={7},
	number={6},
	pages={641},
	year={2025},
	publisher={Springer}
}

@inproceedings{DynPose2025,
	title={Dynpose: Largely improving the efficiency of human pose estimation by a simple dynamic framework},
	author={Xu, Yalong and Zhao, Lin and Gong, Chen and Li, Guangyu and Wang, Di and Wang, Nannan},
	booktitle=CVPR,
	pages={1160--1169},
	year={2025},
	organization={IEEE}
}

@article{PoseNet++,
	title={PoseNet++: A multi-scale and optimized feature extraction network for high-precision human pose estimation},
	author={Lv, Chao and Ma, Geyao},
	journal={Plos one},
	volume={20},
	number={6},
	pages={e0326232},
	year={2025},
	publisher={Public Library of Science, San Francisco, CA USA}
}
\vspace{18mm}
\begin{IEEEbiography}[{\includegraphics[width=1in,height=1.25in,clip,keepaspectratio]{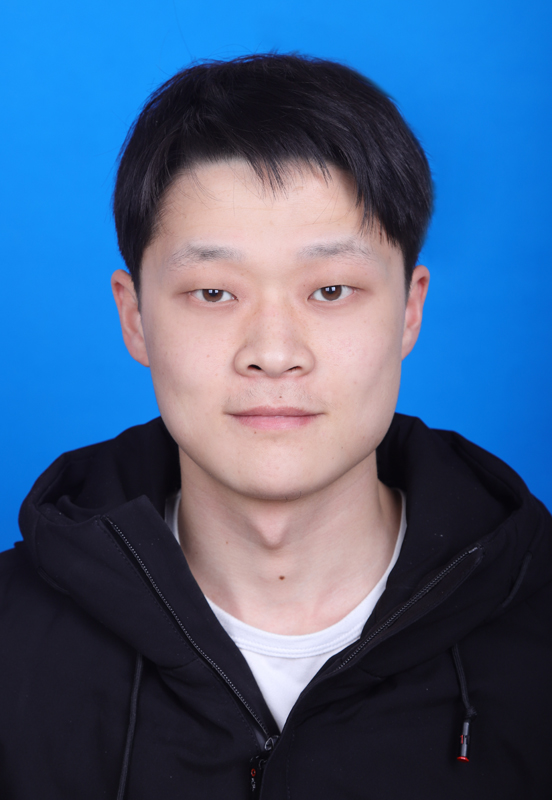}}]{Shibang Liu} received the B.E. degree from East China Jiaotong University, Nanchang, China, in 2019, and the Ph.D. degree from Xidian University, Xi'an, China, in September 2026. His current research interests include computer vision, image/motion generation, image processing, and pattern recognition.
\end{IEEEbiography}

\begin{IEEEbiography}[{\includegraphics[width=0.9in,clip,keepaspectratio]{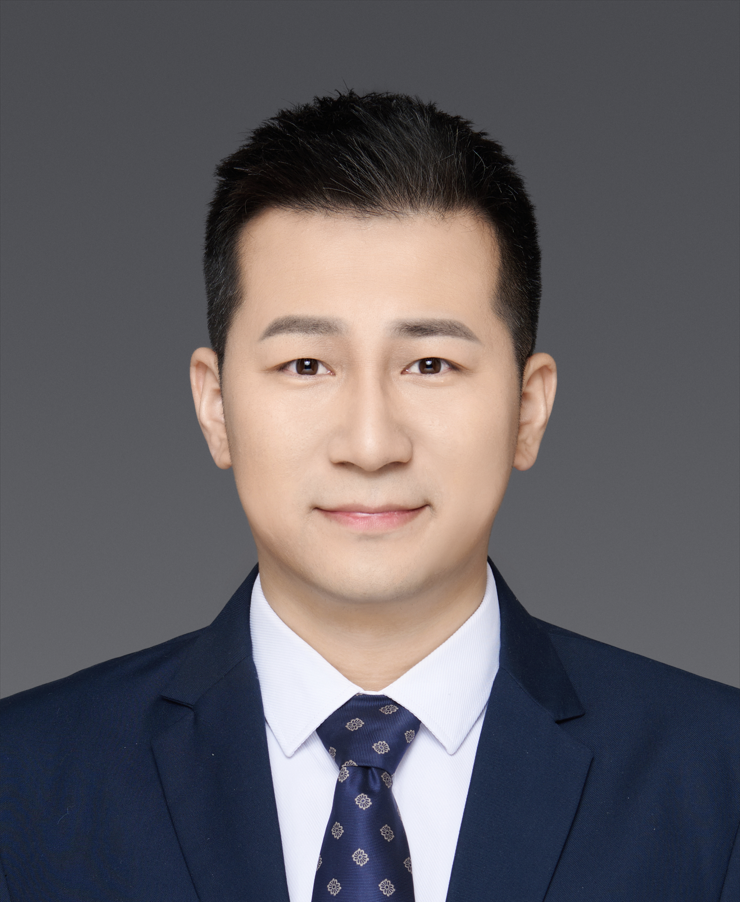}}]
	{Xuyang Li} received the M.S. degree in Measurement, Control Technology and Instrumentation from Xidian University, Xi'an, China, in 2014 and the Ph.D. degree in Computer Science and Technology, from Xidian University in 2023. He is currently a lecturer at the Guangzhou Institute of Technology at Xidian University, Guangdong, China. His primary research focuses on embodied perception in robotics, including 6D pose estimation and multimodal fusion sensing.
	
\end{IEEEbiography}

\begin{IEEEbiography}[{\includegraphics[width=0.9in,clip,keepaspectratio]{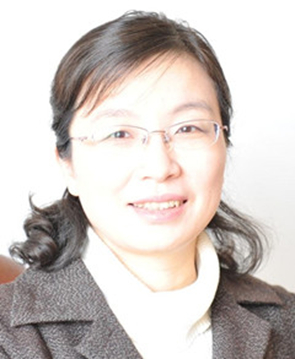}}]{Xuemei Xie} (Senior Member, IEEE) received the M.S. degree in electronic engineering from Xidian University, Xi’an, China, in 1994, and the Ph.D. degree in electrical and electronic engineering from The University of Hong Kong, Hong Kong, in 2004. She is currently a Professor with the School of Artificial Intelligence, Xidian University. She has authored more than 100 academic papers in international and national journals, and international conferences. Her research interests include scene understanding, embodied intelligence robotics, natural language processing, vision-language multimodal learning, and human–computer interaction.
\end{IEEEbiography}

\begin{IEEEbiography}[{\includegraphics[width=0.9in,height=1.25in,clip,keepaspectratio]{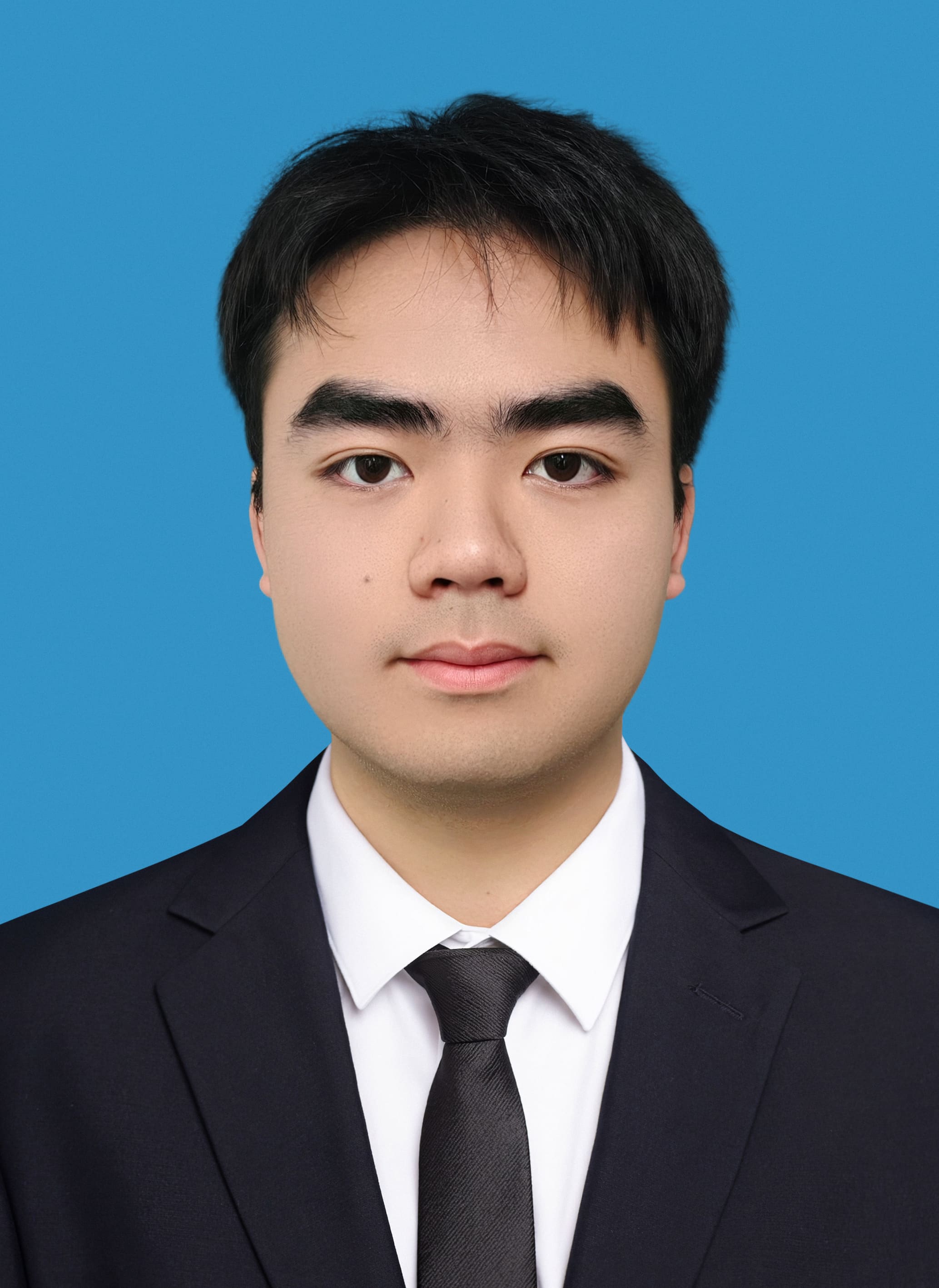}}]{Yi Luo}  received the B.E. degree from East China Jiaotong University, Nanchang, China, in 2019. He is currently pursuing the Ph.D. degree at Xidian University, Xi'an, China. His current research interests include computer vision, image processing, and pattern recognition.
\end{IEEEbiography}

\end{document}